\documentclass[prx,reprint,superscriptaddress,nofootinbib,amsmath,amssymb]{revtex4-2}
\usepackage[T1]{fontenc}
\usepackage{lmodern}
\usepackage{bm}
\usepackage{graphicx}
\usepackage{booktabs}
\usepackage{algpseudocode}
\usepackage{framed}
\usepackage{placeins}
\usepackage{xurl}

\usepackage[colorlinks=true,linkcolor=blue,citecolor=blue,urlcolor=blue]{hyperref}
\usepackage{orcidlink}
\graphicspath{{figures/}}
\newcommand{\dC}{d_C}
\newcommand{\Pstar}{P^{*}}
\newcommand{\tr}{\operatorname{tr}}

\begin{document}
\title{A phase diagram for training and inference in language models}
\author{Zhuo-Yang Song~\orcidlink{0009-0001-9727-7908}}
\email{zhuoyangsong@stu.pku.edu.cn}
\affiliation{School of Physics, Peking University, Beijing 100871, China}
\affiliation{LightRay AI Lab, Beijing, China}

\author{Zeyu Li}
\affiliation{School of Physics, Peking University, Beijing 100871, China}
\affiliation{Center for High Energy Physics, Peking University, Beijing 100871, China}
\affiliation{LightRay AI Lab, Beijing, China}

\author{Qing-Hong Cao}
\email{qinghongcao@pku.edu.cn}
\affiliation{School of Physics, Peking University, Beijing 100871, China}
\affiliation{School of Physics, Zhengzhou University, Zhengzhou 450001, China}
\affiliation{Center for High Energy Physics, Peking University, Beijing 100871, China}

\author{Ming-xing Luo}
\email{mingxingluo@csrc.ac.cn}
\affiliation{Beijing Computational Science Research Center, Beijing 100193, China}

\author{Hua Xing Zhu}
\email{zhuhx@pku.edu.cn}
\affiliation{School of Physics, Peking University, Beijing 100871, China}
\affiliation{Center for High Energy Physics, Peking University, Beijing 100871, China}

\begin{abstract}
A language model is a complex system whose behavior is determined jointly by the Transformer and the corpus, yet how the representation geometry in the Transformer relates to the actual behavior of language models remains open. We analyze how the representation geometry changes along the forward propagation of actual language models and characterize the representation geometry of a document at a given layer by two collective coordinates, document distinguishability and document dimension. On the phase diagram spanned by these collective coordinates, inference is represented as a flow along depth. We find that, for untrained models, depending on their initialization scale, most of the 156 inference flows move toward the collapse and noise regions of the phase diagram. By contrast, the inference flows of 8 trained models from 5 architecture families all enter and reside in the central region of the phase diagram. Describing the evolution of the flows with a discrete Langevin equation, we find that the difference between the inference flows of untrained runs and those of trained models reduces to whether the effective force on the flows in this region is of saddle or stable-node type. On Pythia-1.4B, we interpret the training process as a phase transition. These results give a unified gateway to the relation between representation geometry and language-model behavior.
\end{abstract}

\maketitle

\section{Introduction}\label{sec:intro}

\begin{figure*}
\centering
\includegraphics[width=0.89\textwidth]{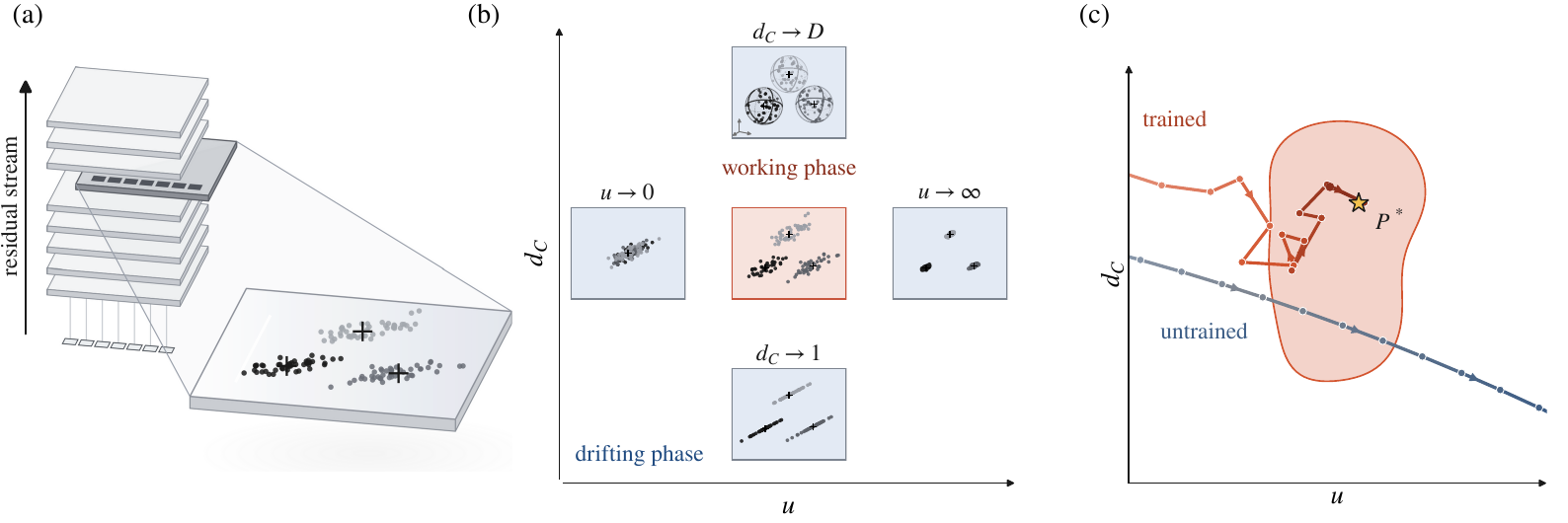}
\caption{Schematic phase diagram of language-model training and inference.
(a) Schematic of residual propagation in a Transformer: the tokens of a document (bottom boxes) enter the network, and their representations propagate layer by layer along the residual stream (arrows); the slab magnifies the representation geometry at a given layer, and the point cloud shows the token representations of the document.
(b) Schematic phase diagram of token representations: the two axes are collective coordinates of the token representations, with document distinguishability $u$ on the horizontal axis and document dimension $\dC$ on the vertical axis. The working phase lies in the middle and the drifting phase forms the outer ring; the insets sketch token-representation point clouds in the middle and near the four limits.
(c) Two trajectories, one point per layer; the trajectory of a trained model resides in the working phase, whereas that of an untrained run drifts away (measurements in Fig.~\ref{fig:coordinates}(d) and Fig.~\ref{fig:diagram}).}
\label{fig:schematic}
\end{figure*}

A language model is a complex system jointly determined by the Transformer architecture and the training corpus~\cite{brown2020language,vaswani2017attention,zhao2023survey,bommasani2021opportunities}.
Observations of its behavior are accumulating rapidly~\cite{kaplan2020scaling,hoffmann2022training,wei2022emergent,nakaishi2024critical,arnold2024phase,sun2025phase,liang2022holistic,srivastava2022beyond}.
All these behaviors can be carried by the token representations, which change layer by layer in the forward propagation~\cite{nostalgebraist2020logit,belrose2023eliciting}; the output is read out from the last-layer representation.
Benchmarks characterize model behavior at the output level~\cite{liang2022holistic,srivastava2022beyond,chang2024survey}, whereas the geometry of internal representations is studied in a growing body of work~\cite{ansuini2019intrinsic,ethayarajh2019contextual,cheng2025emergence}.

Statistical mechanics has long been used to analyze learning in neural networks~\cite{gardner1988space,seung1992statistical,engel2001statistical,carleo2019machine,bahri2020statistical}. Depending on the hyperparameters, forward propagation in randomly initialized neural networks can lie in an ordered or a chaotic phase~\cite{poole2016exponential,schoenholz2017deep,roberts2022principles,cowsik2024geometric}, and the learning process of shallow Transformers has likewise been characterized mechanistically~\cite{nanda2023progress,gopalani2025what}. 
Trained language models, however, are also shaped by the corpus, tokenization and embedding, and the training process. Studies of their representations have shown that layer-wise probes can recover syntactic, semantic, and factual information~\cite{tenney2019bert,hewitt2019structural,liu2021probing}, and that representations across different models converge with scale~\cite{kornblith2019similarity,huh2024platonic}. These observations are expressed in different quantities and settings, which makes them difficult to compare.

\begin{figure*}[htbp]
\centering
\includegraphics[width=\textwidth]{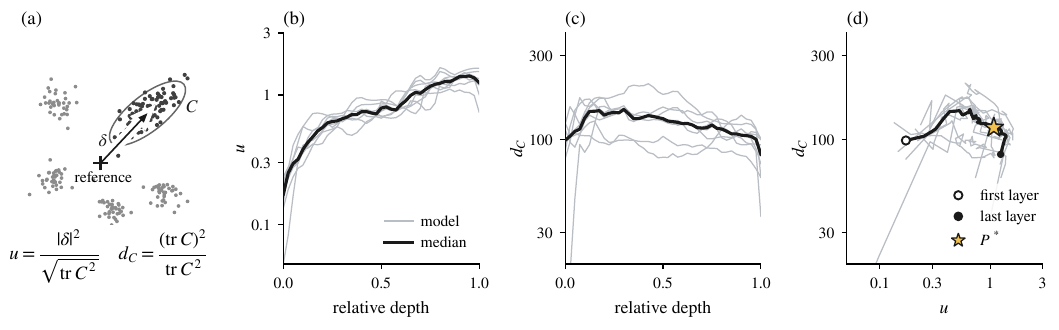}
\caption{Collective coordinates and the inference flows of trained models: $u$ rises with depth, $\dC$ changes little, and the trajectories stop in a small region. (a) Schematic: dark gray points are the token representations of one document $d$ at a given layer; light gray points are those of the other documents; the cross marks the mean of the other documents' token representations, the offset $\delta$ points from the mean of the other documents to the mean of this document's token representations, and the ellipse represents the within-document covariance $C$; the definitions of $u$ and $\dC$ are shown at the bottom.
(b), (c) $u$ and $\dC$ versus relative depth: gray lines are the document-median trajectories of individual models (medians over 96 documents), and the black line is the median of the 8 trajectories at each relative depth; the vertical axes are logarithmic.
(d) The same trajectories on the $(u,\dC)$ plane, with both axes logarithmic; the open circle and the filled dot mark the first and last layers of the black line, and the star marks $\Pstar=(1.08,117)$, where the inference flows of trained models end up (Sec.~\ref{sec:diagram}).}
\label{fig:coordinates}
\end{figure*}

To compare them, we view the Transformer, through its residual stream~\cite{elhage2021mathematical}, as a discrete dynamical system along depth~\cite{he2016deep,e2017proposal,haber2017stable,chen2018neural,lu2019understanding,sander2022residual,geshkovski2023mathematical} and study how the representation geometry changes in its forward propagation. Figure~\ref{fig:schematic} shows this framework. The token representations of a document propagate layer by layer along the residual stream (Fig.~\ref{fig:schematic}(a)). We characterize the representation geometry of a document at a given layer by two collective coordinates, the \textbf{document distinguishability} $u$ and the \textbf{document dimension} $\dC$ (Fig.~\ref{fig:coordinates}(a)). 
The representation geometry at each layer is a point on the phase diagram (Fig.~\ref{fig:schematic}(b)). 
The points of successive layers form a trajectory, and a model's trajectories over all documents make up its inference flow. 
We describe the evolution of the trajectories with a discrete Langevin equation~\cite{vankampen2007stochastic}.

Since the output of a language model is read out from the last-layer representation, the phase diagram is partitioned by where the trajectories end up. 
We call the central region where inference flows of trained models reside the working phase and denote its representative point by $\Pstar$. The surrounding region toward which inference flows of untrained runs drift is called the drifting phase.
The boundary between the two phases can be drawn directly from measurements.
The training step serves as the control parameter that determines where the trajectories end up. 
As training proceeds, the inference flow near $\Pstar$ changes from non-contracting to contracting (Fig.~\ref{fig:transition}). 
In this way, we distinguish how inference flows behave before and after training on the phase diagram spanned by the collective coordinates.
To our knowledge, this is the first phase diagram of language models constructed from internal measurements of representations.

On this phase diagram, untrained and trained networks fall into two distinct phases according to where their inference flows end up (Fig.~\ref{fig:diagram}).
Depending on the initialization scale, inference flows of untrained runs drift toward either the collapse region or the noise region of the drifting phase (Fig.~\ref{fig:diagram}(a)). By contrast, the inference flows of trained models, which come from different architecture families and differ in training scheme, all enter and reside in the working phase (Fig.~\ref{fig:diagram}(b)) and differ among themselves far less than those of untrained networks do. 
In the measured models, this distinction does not depend on the specific training or initialization scheme. 
We can therefore distinguish trained from untrained models by examining how their representations propagate through the network, without decoding. Residing in the working phase is a statistical property shared by actual language models. The definitions of the collective coordinates also allow many existing observations to be mapped directly onto this diagram (Sec.~\ref{sec:discussion} and Appendix~\ref{app:dictionary}).

Section~\ref{sec:coordinates} defines the collective coordinates. Section~\ref{sec:diagram} presents the phase diagram and explains how the fates of inference flows distinguish the two phases. Section~\ref{sec:phases} describes the flow dynamics using a discrete Langevin equation and relates the distinction between the two phases to the type of effective force at $\Pstar$. Section~\ref{sec:transition} examines the phase transition during the training of Pythia-1.4B. Finally, Section~\ref{sec:discussion} discusses the implications for the relation between representation geometry and model behavior, together with open questions.

\section{Collective coordinates}\label{sec:coordinates}

\subsection{Definitions}\label{sec:coordinates:def}

To characterize the effective behavior of networks before and after training, we use geometric quantities derived from their representations as collective coordinates, rather than hyperparameters associated with training or architecture.
Both randomly initialized Transformers and trained language models can thus be mapped onto the same phase diagram using their representations alone, regardless of the underlying representation space.

We use the document as the unit of analysis because attention incorporates document-level features, such as topic, references to the same entities~\cite{clark2019what}, and style, into the token representations.
At a fixed layer, let $h\in\mathbb{R}^D$ denote the representation of a token in a document indexed by $d$, and let $\mu_d$ denote the mean representation over all tokens in that document. Here, $D$ is the embedding-space dimension of the Transformer.
The reference $\mu_{\mathrm{ref}}$ is the mean of all token representations of the other 95 documents in the corpus of 96 long documents, and we write (Fig.~\ref{fig:coordinates}(a))
\begin{equation}
\delta=\mu_d-\mu_{\mathrm{ref}},\qquad
C=\bigl\langle (h-\mu_d)(h-\mu_d)^{\mathsf T}\bigr\rangle_d ,
\label{eq:moments}
\end{equation}
where $\langle\cdot\rangle_d$ denotes the average over tokens in the document. The mean shift common to all documents cancels in $\delta$.

Since representation bases and scales differ across networks, the collective coordinates are two combinations of $(\delta,C)$ that are invariant under rotation and overall rescaling,
\begin{equation}
u=\frac{|\delta|^2}{\sqrt{\tr C^2}},\qquad
\dC=\frac{(\tr C)^2}{\tr C^2}.
\label{eq:coords}
\end{equation}
The document distinguishability $u$ is the squared offset of the document mean, $|\delta|^2$, normalized by the scale of the within-document variation, $\sqrt{\tr C^2}$.
A larger $u$ makes it easier to identify a token's source document from its representation alone (Appendix~\ref{app:coordinates:readout}).
The document dimension $\dC$ is the participation ratio of $C$~\cite{bell1970atomic,litwinkumar2017optimal} and measures the effective number of directions along which token representations vary within a document.
If the variation is equally distributed over $k$ directions, then $\dC=k$. For a given document and layer, the pair $(u,\dC)$ defines its state on the phase diagram. Since both coordinates are positive ratios that span several orders of magnitude (Fig.~\ref{fig:diagram}), we describe state changes in relative terms. Accordingly, we use the logarithmic coordinates $\log_{10}u$ and $\log_{10}\dC$ (in units of dex) for both plotting and dynamical analysis. Changing the logarithm base introduces only an overall rescaling of these coordinates.

\subsection{Measurement}\label{sec:coordinates:measurement}

Appendix~\ref{app:data} describes the corpus, the 8 trained models, and the initialization schemes used for the untrained networks. For each document, we perform a single forward propagation using its first 16\,384 tokens, truncated as needed to fit the model's context length.
We first discard the first few tokens of each document and any remaining tokens whose residual-stream norms greatly exceed the document median~\cite{xiao2024efficient,sun2024massive} (Appendix~\ref{app:data:measurement}). We then measure the residual stream at the outputs of blocks $1$ through $L-1$, where $L$ is the total number of Transformer blocks. We exclude the embedding layer, which maps input tokens to vectors, and the final block, whose output is used for next-token prediction, because both are closely tied to the vocabulary. We define the relative depth of block $t$ as $(t-1)/(L-2)$.

In a single forward propagation, a document's states at successive layers trace out a trajectory on the phase diagram. The trajectories over all selected documents together constitute the model's inference flow. Connecting the per-layer medians over the 96 documents gives the model's document-median trajectory. During residual propagation, each layer adds an increment to the representations, and adjacent layers have very similar states. We therefore treat the inference flow as a quasi-continuous dynamical system~\cite{lu2019understanding,sander2022residual}.

In trained models, $u$ rises rapidly in the early layers and slowly thereafter, while $\dC$ shows a slight initial increase followed by a gradual decline (Fig.~\ref{fig:coordinates}(b)(c)). The trajectories of all 8 models lie close together and end in the same small region of the phase diagram (Fig.~\ref{fig:coordinates}(d)).
\section{The fate of the flow determines the phase}\label{sec:diagram}

\begin{figure*}
\centering
\includegraphics[width=\textwidth]{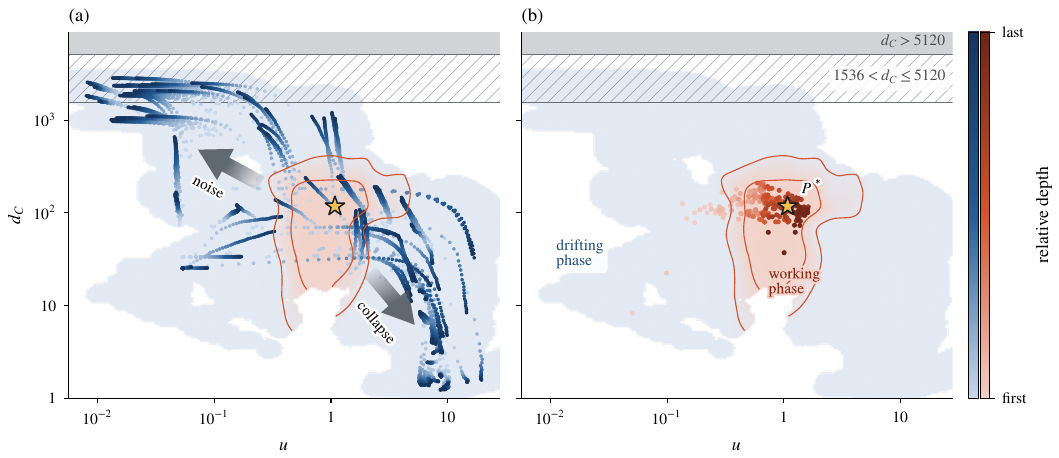}
\caption{Fates of inference flows before and after training: most inference flows of untrained runs go to the collapse region or the noise region, whereas all inference flows of trained models reside in the working phase.
(a) 156 untrained runs; (b) 8 trained models. Both axes, $u$ and $\dC$, are logarithmic. Each point is the position of one network's document-median trajectory at one layer; color runs from light to dark with relative depth from the first to the last layer. The background shows the measured phase: light red is the working phase, light blue is the drifting phase, and the band between the two red lines is the coexistence band of the two phases (working-phase fraction 0.16--0.84; the red lines are contours of the smoothed fraction, see Appendix~\ref{app:sensitivity:phase}); white areas have no deep-layer data. The star marks $\Pstar=(1.08,117)$, where inference flows of trained models end up.
The two arrows in (a), labeled ``collapse'' and ``noise'', indicate the fates of inference flows of untrained runs, which depend on the initialization scale; they point from $\Pstar$ to the collapse region at large $u$ and small $\dC$ and to the noise region at small $u$ and large $\dC$ (Appendix~\ref{app:data:fates}). The embedding-space dimension $D$ ranges from 1536 to 5120 for the measured networks. Because $\dC\le D$, the solid gray area is unreachable for all networks, and the hatched band is unreachable for networks whose $D$ is smaller than the local $\dC$.}
\label{fig:diagram}
\end{figure*}

We construct 156 untrained runs by replacing the parameters inside the Transformer blocks of the trained models with random samples, using several initialization schemes and scales ranging from 0.005 to 1 (Appendix~\ref{app:data:untrained}). Each untrained run is specified by an architecture, an initialization scheme, a scale, and a random seed. The trained models are evaluated with their learned weights held fixed.

Figure~\ref{fig:diagram} compares the inference flows of untrained runs (a) and trained models (b). As depth increases, inference flows of untrained runs generally continue to drift toward opposite sides of the outer ring. Inference flows of trained models, by contrast, enter a small central region within the first few layers and remain there.

The fates of inference flows of untrained runs depend on the initialization scale. At small scales, most flows follow the lower-right arrow in Fig.~\ref{fig:diagram}(a) toward the collapse region (large $u$, small $\dC$). At large scales, most follow the upper-left arrow toward the noise region (small $u$, large $\dC$). Near the transition between these two regimes, the fates are mixed (Appendix~\ref{app:data:fates}). These two outcomes are consistent with rank collapse~\cite{dong2021attention,noci2022signal} and entropy collapse~\cite{giorlandino2026two} in random Transformers, respectively.

We assume that the fate of an inference flow determines its phase. Because the networks have finite depth, we use their late-half states as proxies for these fates. We pool these states across all documents and networks into two point clouds, one untrained and one trained, and use them to delineate the phase boundary (Appendix~\ref{app:sensitivity:phase}). The central region occupied by inference flows of trained models is the working phase (light red in Fig.~\ref{fig:diagram}), while the outer ring occupied by inference flows of untrained runs is the drifting phase (light blue).

The two point clouds overlap in a coexistence band bounded by the two red lines. The width of this band may reflect finite-depth effects: inference flows are still evolving at the measured depths, so the band is expected to narrow as depth increases. The distinction between the two phases is empirical, based on the regions occupied by inference flows in the deep layers before and after training. The location of the working phase may vary slightly with the choice of corpora and models.

To mark where the inference flows of trained models settle, we take medians over documents, layers, and models in turn, each separately in $\log_{10}u$ and $\log_{10}\dC$. These give, respectively, a model's document-median trajectory, its late-half median position (over the layers with relative depth $\ge 1/2$), and $\Pstar$, the star in Fig.~\ref{fig:coordinates}(d).

The 8 trained models represent 5 architecture families, with depths ranging from 28 to 48 blocks and embedding-space dimensions from 1536 to 5120. They were trained by 4 institutions using different corpora and training recipes, yet their inference flows all reside in the working phase (Fig.~\ref{fig:diagram}(b)). Their document-median trajectories enter the working phase (the light red region in Fig.~\ref{fig:diagram}) at a median relative depth of 0.11 and remain there at every subsequent layer.

Trained models also show much less variation in their deep-layer positions than untrained runs. The dispersion of the late-half median positions, measured by their root-mean-square distance from the group mean, is 0.085~dex for the 8 trained models, compared with 0.3 to 1.4~dex for groups of untrained runs sharing an initialization scheme but differing in architecture (Appendix~\ref{app:data:fates}). These results suggest that residing in the working phase is a statistical property shared by actual language models, consistent with the observed convergence of representations across models~\cite{kornblith2019similarity,huh2024platonic}.

The drifting phase can be understood in terms of four limits. As $u\to0$, documents become indistinguishable; as $u\to\infty$, the within-document variation vanishes relative to the mean offset $\delta$. As $\dC\to1$, the within-document variation is confined to a single direction; as $\dC\to D$, it becomes isotropic across all $D$ directions. The representative point $\Pstar$ of the working phase lies between these limits: documents remain distinguishable without their representations collapsing, while the within-document variation spans many effective directions without becoming isotropic across the full space.

Some inference flows of untrained runs pass through the working phase without residing there (Fig.~\ref{fig:diagram}(a)), whereas inference flows of trained models reaching the same region remain there. Inference flows of untrained runs slow down progressively with depth, but drift remains observable up to the deepest measured layers. At finite depth, position alone is therefore insufficient to determine a flow's phase or explain why inference flows evolve toward different phases. Both questions require a dynamical description.
\section{Dynamical description of the phases}\label{sec:phases}

\begin{figure*}
\centering
\includegraphics[width=\textwidth]{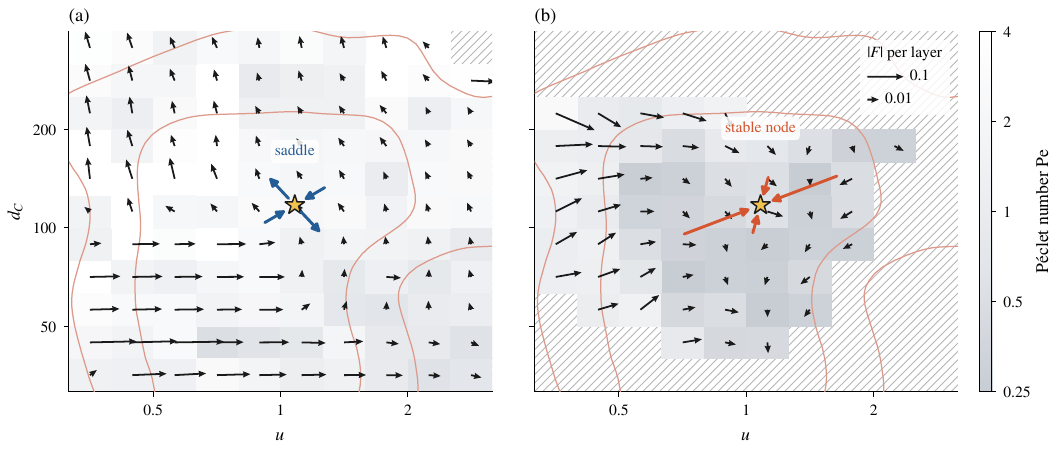}
\caption{Type of the effective force at $\Pstar$: saddle for the untrained group, stable node for the trained group. (a) 156 untrained runs; (b) 8 trained models. Both panels magnify the range $0.3\le u\le3$, $30\le\dC\le400$ of Fig.~\ref{fig:diagram}; faint red lines are the edges of the coexistence band in Fig.~\ref{fig:diagram}; cells are 0.1~dex on a side. Black arrows show the effective force $F$ in each cell, with length proportional to $\sqrt{|F|}$. The background is the Péclet number $\mathrm{Pe}$, the ratio of effective force to fluctuation, on a logarithmic color scale; darker means stronger dominance of fluctuation; hatched cells have insufficient data. The star marks $\Pstar=(1.08,117)$, where inference flows of trained models end up (Sec.~\ref{sec:diagram}); the effective forces of both groups are linearized there. Both axes are logarithmic. Colored arrows show the two eigendirections of $J$ in Eq.~\eqref{eq:linear}, with length proportional to $\sqrt{|\lambda|}$, where $\lambda$ is the corresponding eigenvalue; arrows pointing inward indicate restoring and arrows pointing outward indicate diverging.}
\label{fig:phases}
\end{figure*}

To understand how inference flows evolve toward different phases, we examine their dynamics near $\Pstar$. We assume that an inference flow can be described by an equation of motion with a small number of parameters, and that this equation determines its fate. We further assume, separately for untrained networks and trained models, that layers acting on the same state have similar effects. Their action can therefore be described by a common equation of motion expressed solely in collective coordinates, with no explicit dependence on the layer index or model parameters. This assumption allows us to pool states across networks and depths within each group for joint estimation on the phase diagram.

We denote the position on the phase diagram and the displacement of each layer as
\begin{equation}
x_t=\begin{pmatrix}\log_{10}u_t\\ \log_{10}d_{C,t}\end{pmatrix},\qquad v_t=x_t-x_{t-1} ,
\label{eq:xv}
\end{equation}
where $t$ is the layer index. If the next step were determined by the state alone, the displacement at each layer would be a function of position only, and the motion would be quasistatic. In a residual network, however, the displacement between layers is determined by the full representation, and degrees of freedom outside the collective-coordinate projection $(u,\dC)$ may affect the next step. To model this influence, we attribute it to the displacement of the previous layer and to a fluctuation term,
\begin{equation}
v_{t+1}=\alpha\,v_t+F(x_t)+\eta_t ,
\label{eq:langevin}
\end{equation}
i.e., a discrete Langevin equation, which describes how the current state determines the displacement produced by the next layer. The effective force $F$ is the mean part of the displacement determined by the current state alone. Different layers usually sit at different positions, so differences in what the layers do are mostly reflected in position; $F$ depends only on the state and not explicitly on the layer index, just as the force on a particle in a stationary force field is set by its position rather than by time. Degrees of freedom outside the collective-coordinate projection enter the equation in two parts. The slower-varying ones let inference flows passing through the same state retain different tendencies of motion; these tendencies are carried to the next layer by $\alpha v_t$, formally like inertia, and $\alpha$ is called the inertia coefficient. In this simplest form of the equation, the remaining degrees of freedom are merged into the fluctuation $\eta_t$, together with the remaining layer-to-layer differences at the same position. The terms $\alpha v_t$ and $\eta_t$ correspond to the memory term and the fluctuation in the Mori--Zwanzig projection-operator formalism~\cite{zwanzig1961memory,mori1965transport}; when $\alpha\approx0$, the memory term nearly vanishes and the degrees of freedom outside the projection enter only as unbiased fluctuation. When $\alpha>0$, the direction of the next layer's step varies with the previous layer's displacement, even for inference flows passing through the same position.

We pool the inference flows of the 156 untrained runs into the untrained group and those of the 8 trained models into the trained group (criterion for contraction, estimation method, and sensitivity analysis in Appendices~\ref{app:linear}--\ref{app:sensitivity}). For each group, one $\alpha$ is fitted over the whole phase diagram, and $F$ is estimated cell by cell~\cite{friedrich2011approaching} on a uniform grid within the fixed fit range $0.3\le u\le3$, $30\le\dC\le400$ and linearized at $\Pstar$,
\begin{equation}
F(x)\approx a+J\,(x-\Pstar),
\label{eq:linear}
\end{equation}
where $\Pstar$ (Sec.~\ref{sec:diagram}) is expressed in the coordinates of Eq.~\eqref{eq:xv}; $a$ is the constant term, and $J$ is the linear response matrix of the effective force at $\Pstar$. The eigenvalues of $J$ determine whether inference flows passing near $\Pstar$ contract.

The inference flows of the two groups show two distinct patterns. The untrained group has a positive inertia coefficient, $\alpha=0.287^{+0.041}_{-0.030}$ (all errors in this section are 16\%--84\% bootstrap~\cite{efron1979bootstrap,efron1994introduction} intervals, Appendix~\ref{app:estimation:bootstrap}). Its effective force at $\Pstar$ is a saddle (Fig.~\ref{fig:phases}(a)), restoring in one direction and diverging in the other; along the diverging direction, flows separate toward the noise region and the collapse region, which are marked by the two arrows in Fig.~\ref{fig:diagram}(a). The trained group has an inertia coefficient of about 0, $\alpha=-0.001^{+0.033}_{-0.021}$, and its motion is quasistatic. Its effective force at $\Pstar$ is a stable node (Fig.~\ref{fig:phases}(b)), restoring along both eigendirections.

With $s_i$ denoting the standard deviation of the fluctuation $\eta_t$ along the $i$th coordinate within a cell (Appendix~\ref{app:estimation:peclet}), the Péclet number of the cell, the ratio of the effective force to the fluctuation, is
\begin{equation}
\mathrm{Pe}=\Bigl[\sum_i \bigl(F_i/s_i\bigr)^2\Bigr]^{1/2} .
\label{eq:peclet}
\end{equation}
Within the fit range, the median $\mathrm{Pe}$ (background in Fig.~\ref{fig:phases}) is 1.99 for the untrained group and 0.49 for the trained group. These results show that at each layer the effective force mainly drives inference flows of untrained runs in directed motion, whereas the single-layer motion of inference flows of trained models is dominated by fluctuations; their effective force is weak, and they move slowly overall.

Substituting Eq.~\eqref{eq:linear} into Eq.~\eqref{eq:langevin} and neglecting the fluctuation term, we find that the deviation $(\delta x,\delta v)$ of inference flows from the average of the flows near $\Pstar$ propagates layer by layer as
\begin{equation}
\begin{pmatrix}\delta x\\ \delta v\end{pmatrix}_{t+1}=M\begin{pmatrix}\delta x\\ \delta v\end{pmatrix}_{t},\qquad
M=\begin{pmatrix}I+J & \alpha I\\ J & \alpha I\end{pmatrix},
\label{eq:propagator}
\end{equation}
where $I$ is the $2\times2$ identity matrix. The contraction rate per layer is $-\ln\rho$, where $\rho$ is the spectral radius of $M$~\cite{elaydi2005introduction}. When $-\ln\rho>0$, the deviation is multiplied by $\rho<1$ per layer and the inference flows contract toward one another; otherwise they do not. At $\Pstar$, $-\ln\rho=-0.022^{+0.003}_{-0.008}$ for the untrained group and $-\ln\rho=0.011^{+0.019}_{-0.006}$ for the trained group. The inference flows of the untrained group, which do not contract, leave $\Pstar$ and drift along the diverging direction of the saddle toward opposite sides of the outer ring.

Therefore, the distinction between the two phases observed in Sec.~\ref{sec:diagram} corresponds to the type of the effective force at $\Pstar$. Inference flows of untrained runs do not contract at $\Pstar$ and are in the drifting phase; inference flows of trained models contract at $\Pstar$ and are in the working phase. The sign of $-\ln\rho$ distinguishes the two phases. We also observe that the inertia coefficient is positive for the untrained group and about 0 for the trained group. Section~\ref{sec:transition} uses the contraction rate and the inertia coefficient to follow the transition between the two phases during the training of a single network.

\section{Emergence of phases during training}\label{sec:transition}

\begin{figure*}
\centering
\includegraphics[width=\textwidth]{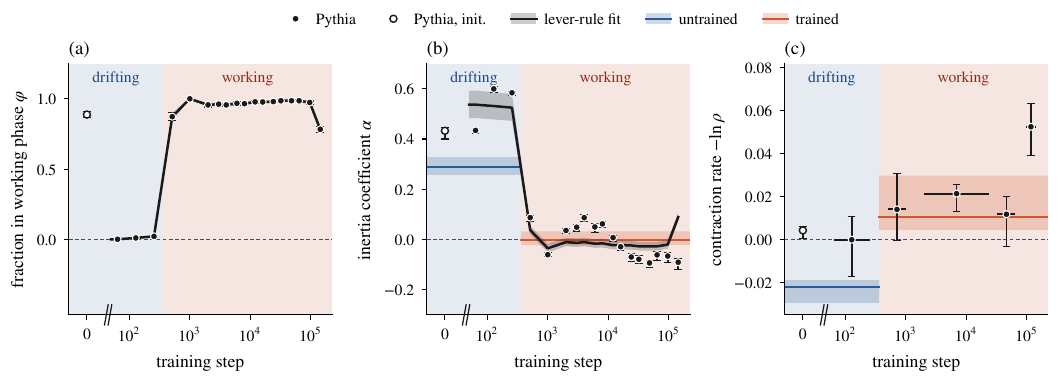}
\caption{Two phases in the training of Pythia-1.4B: the fraction of deep-layer states in the working phase and the inertia coefficient $\alpha$ jump simultaneously between steps 256 and 512, after which $-\ln\rho$ is positive. The horizontal axis is the training step, on a logarithmic scale; background shading marks the two phases, with step 0 assigned to the drifting phase (Appendix~\ref{app:sensitivity:criterion}). All vertical error bars are bootstrap 16\%--84\% intervals. (a) Fraction $\varphi$ of each checkpoint's late-half-layer states (96 documents) that fall in the working phase of Fig.~\ref{fig:diagram}; the line connects the checkpoints, and the open point is step 0, shown separately across the axis break. (b) Inertia coefficient $\alpha$ fitted separately for each checkpoint; the black line is the lever-rule fit~\cite{callen1985thermodynamics} evaluated with $\varphi$ from (a), with its bootstrap 16\%--84\% band (Appendix~\ref{app:sensitivity:lever}). (c) Contraction rate $-\ln\rho$ fitted jointly over the checkpoints in each training-step bin; uncapped horizontal bars give the step range of the checkpoints in the bin. In (c), the black points use the late-half median position $\Pstar_{\rm Py}=(0.63,27)$ of the final Pythia-1.4B checkpoint as the linearization point; the blue and red lines are the values for the untrained group (156 runs) and the trained group (8 models) of Sec.~\ref{sec:phases} at $\Pstar=(1.08,117)$, with bands showing the same 16\%--84\% intervals. These lines are for reference only and are not compared value by value with the black points, because the two differ in the models measured, the input length, and the linearization point.}
\label{fig:transition}
\end{figure*}

With the training step as the control parameter, we measure the fraction of deep-layer states in the working phase and the inertia coefficient $\alpha$ at each of 19 checkpoints of Pythia-1.4B~\cite{biderman2023pythia}, spanning initialization to the end of training and spaced roughly uniformly in the logarithm of the step (Appendix~\ref{app:data:pythia}). Because the contraction rate $-\ln\rho$ requires sufficient data near the linearization point, we fit it in training-step bins to reduce the estimation error of a single checkpoint (Appendix~\ref{app:sensitivity:bins}). We thus observe how the phase of the inference flow, the inertia coefficient, and the contraction rate change during training.

Figure~\ref{fig:transition} compares these three quantities over the training of one network with their values for the two groups of Sec.~\ref{sec:phases}. Between steps 256 and 512, the fraction of deep-layer states in the working phase jumps from almost zero to the vast majority; the boundary between the two phases lies in this interval. In the same interval the inertia coefficient drops abruptly to near 0, a change far larger than its variation between later checkpoints. Weighting the inertia coefficients of the two phases by the working-phase fraction (the lever rule) reproduces this drop (Fig.~\ref{fig:transition}(b), Appendix~\ref{app:sensitivity:lever}). The contraction rate is about 0 in the drifting phase and turns positive after the flow enters the working phase, without such a jump.

At step 0 the deep-layer states also fall in the working phase, yet the inertia coefficient matches that of the drifting phase; the inference flow is still drifting and only happens to pass through the working phase at finite depth (Appendix~\ref{app:sensitivity:criterion}). This is precisely the situation noted in Sec.~\ref{sec:diagram}, in which position alone cannot determine the phase and a dynamical description is needed. These observations agree with the results of Sec.~\ref{sec:phases}. Once training brings the inference flow into the working phase, the contraction rate becomes positive, while the inertia coefficient jumps to near 0. We interpret this transition between the two phases during training as a phase transition, consistent with abrupt changes reported during training, such as the onset of generalization in grokking~\cite{power2022grokking} and the formation of induction heads~\cite{olsson2022context}.

\section{Discussion}\label{sec:discussion}

In this work, we show that residual propagation in a Transformer can be described on a phase diagram defined by collective coordinates derived from its representations. A language model can be mapped onto this diagram using a single forward propagation over a common set of documents. The residual dynamics exhibit two phases depending on training progress, with an empirically determined boundary between them (Fig.~\ref{fig:diagram}).

In the 8 trained models, layer-to-layer propagation depends only on the preceding state, and inference flows enter and remain in the working phase in the deep layers. In insufficiently trained models, propagation also depends on the preceding few layers, suggesting that degrees of freedom outside the collective-coordinate projection influence the dynamics. Their inference flows tend to drift toward the collapse or noise region in the deep layers. This distinction holds across the training and initialization schemes examined.

In dynamical terms, the effective force near $\Pstar$ has a saddle structure in the drifting phase and a stable-node structure in the working phase (Fig.~\ref{fig:phases}). In Pythia-1.4B, training drives a phase transition from the drifting phase to the working phase. Across this transition, the contraction rate becomes positive, while the inertia coefficient jumps to its working-phase value (Fig.~\ref{fig:transition}).

This phase diagram provides a new perspective on model training and inference and serves as a dictionary of existing observations (Appendix~\ref{app:dictionary}). Observations of representation geometry made in different lines of work and with different methods become positions or changes in the same coordinates, so they can be compared with one another and related to language model behavior.

Rank collapse~\cite{dong2021attention,noci2022signal}, neural collapse with documents treated as classes~\cite{papyan2020prevalence}, and the merging of token representations into a single cluster in self-attention dynamics~\cite{geshkovski2023emergence,geshkovski2023mathematical} all correspond to the $u\to\infty$ limit of the drifting phase (Sec.~\ref{sec:diagram}). In random Transformers, rank collapse and entropy collapse occur at different attention initialization scales~\cite{giorlandino2026two}. These two outcomes are consistent with inference flows of untrained runs drifting toward the collapse and noise regions of the drifting phase, respectively (Fig.~\ref{fig:diagram}(a)). Random networks are most easily trained when initialized near the boundary between these two outcomes~\cite{schoenholz2017deep,bahri2020statistical,roberts2022principles}. Near this boundary, the deep-layer states already lie close to $\Pstar$ (Appendix~\ref{app:data:fates}), so training need only modify the effective force to keep the inference flows there, potentially requiring smaller changes to the weights. The observation that some intermediate and deep layers can be removed or reordered with little effect on performance~\cite{veit2016residual,gromov2025unreasonable,men2025shortgpt,lad2024remarkable} is consistent with inference flows of trained models residing in the working phase (Fig.~\ref{fig:diagram}(b)).

The optimal choice of collective coordinates and the limits of their descriptive power remain open questions. The two coordinates used here have a geometric interpretation and are also linked to linear-readout performance (Appendix~\ref{app:coordinates:readout}). Specifically, document identification accuracy measures how reliably a token's source document can be identified from its representation alone. This accuracy increases with $E=u/(\sqrt{\dC}+u)$, which approximates the average cosine similarity between token representations from the same document. Along inference flows of trained models, this accuracy increases with depth, consistent with the observation that layer-wise probes become more accurate in deeper layers~\cite{alain2017understanding}.

Observations made during training can also be compared with changes in inference-flow trajectories on the phase diagram. The initial decrease and subsequent increase in effective rank~\cite{roy2007effective} during pretraining~\cite{li2025tracing} resemble a pattern in $\dC$ over training. Likewise, the early rise in probe accuracy for linguistic knowledge~\cite{liu2021probing} resembles an early increase in $E$ and hence in document identification accuracy. The evolution of the inference flow during Pythia-1.4B training is consistent with these observations (Appendix~\ref{app:dictionary}).

The phase diagram also raises several open questions. First, describing training itself as dynamics on the diagram could deepen our understanding of both training and inference in language models. Second, investigating how the corpus, training procedures, and pretraining and post-training methods shape inference flows could clarify the underlying meaning of the phase diagram. Finally, the choice of collective coordinates determines both the descriptive power and the level of detail of the diagram. Higher-order collective coordinates may capture more complex training and inference dynamics, and a natural next step is to connect the underlying computation to macroscopic regularities of training and generation~\cite{kaplan2020scaling,hoffmann2022training,song2026detailed}.

\FloatBarrier
\section*{Code and data}
The code for the measurement, reduction, analysis, and figures is available at \url{https://github.com/SonnyNondegeneracy/llm-phase-diagram}; the data needed to reproduce all results are available at \url{https://huggingface.co/datasets/Nondegeneracy/llm-phase-diagram-data}. The code is released under the MIT License and the data under the CC BY 4.0 license.

\begin{acknowledgments}
This work is supported by the National Natural Science Foundation of China under Grants No. 12425505, No. 12235001, and No. U2230402, and by the Fundamental and Interdisciplinary Disciplines Breakthrough Plan of the Ministry of Education of China (Grant No. JYB2025XDXM204). AI tools were used to assist with code development and language editing; the authors take full responsibility for the content of this paper.
\end{acknowledgments}

\bibliography{refs}

\onecolumngrid\clearpage\twocolumngrid
\appendix
\section{Phase-diagram dictionary of existing observations}\label{app:dictionary}

The quantities and perspectives used in the phase diagram come from many existing works, and the phenomena observed on it in turn connect many results across the literature. Table~\ref{tab:dictionary} lists each quantity, perspective, and phenomenon with its description in the literature and its sources.

\begingroup
\newcommand{\nm}[1]{\citeauthor{#1}~\cite{#1}}
\newcommand{\dgroup}[1]{\multicolumn{3}{@{}l}{\textbf{#1}}\tabularnewline[2pt]}
\begin{table*}
\footnotesize\setlength{\tabcolsep}{4pt}\renewcommand{\arraystretch}{1.2}
\caption{Phase-diagram dictionary of existing observations. The ``Foundations'' group lists the quantities and perspectives used in this paper and their sources; the ``Phenomena'' group lists the phenomena observed in this paper and the works connected to them. When labels instead of documents serve as classes, $\delta$ is the difference of class means and $C$ the within-class covariance.}\label{tab:dictionary}
\begin{tabular}{@{}p{0.25\textwidth}p{0.40\textwidth}p{0.31\textwidth}@{}}
\toprule
Quantity or phenomenon on the phase diagram & Description in the literature & Sources \tabularnewline\midrule
\dgroup{Foundations}
\raggedright Inference flow: a discrete dynamical system along depth
& \raggedright Residual networks are discretizations of ordinary differential equations
& \raggedright \nm{e2017proposal}; \nm{haber2017stable}; \nm{chen2018neural}; \nm{lu2019understanding} \tabularnewline\addlinespace
\raggedright $\dC$: participation ratio of the within-document covariance
& \raggedright Participation ratio; effective rank and intrinsic dimension measure dimensionality in other ways
& \raggedright \nm{roy2007effective}; \nm{facco2017estimating}; \nm{ansuini2019intrinsic} \tabularnewline\addlinespace
\raggedright $u$: ratio of the between-document offset to the within-document variation
& \raggedright Ratio of between-class to within-class scatter
& \raggedright \nm{papyan2020prevalence} \tabularnewline\addlinespace
\raggedright $E=u/(\sqrt{\dC}+u)$: approximately the average cosine similarity between the representations of two tokens from the same document; document identification accuracy rises with it (Appendix~\ref{app:coordinates:readout})
& \raggedright Accuracy of linear probes
& \raggedright \nm{alain2017understanding}; \nm{belinkov2022probing} \tabularnewline\addlinespace
\raggedright $\delta$: offset of the document mean relative to the reference
& \raggedright Difference-of-means directions: linear directions of concepts such as truth and gender
& \raggedright \nm{park2024linear}; \nm{marks2024geometry} \tabularnewline\addlinespace
\raggedright The mean shared by all documents, removed in $\delta$
& \raggedright Anisotropy: high cosine similarity between representations in upper layers
& \raggedright \nm{ethayarajh2019contextual}; \nm{gao2019representation}; \nm{mu2018all} \tabularnewline
\addlinespace
\dgroup{Phenomena}
\raggedright Inference flows of trained models: $u$ rises with depth, and $\dC$ first rises slightly and then falls slightly (Fig.~\ref{fig:coordinates})
& \raggedright Linear-probe accuracy rises with depth; intrinsic dimension first rises and then falls with depth
& \raggedright \nm{alain2017understanding}; \nm{ansuini2019intrinsic}; \nm{valeriani2023geometry}; \nm{cheng2025emergence} \tabularnewline\addlinespace
\raggedright Inference flows of trained models reside in the working phase (Fig.~\ref{fig:diagram}(b))
& \raggedright Layer redundancy: some intermediate and deep layers can be deleted or reordered with little change in performance
& \raggedright \nm{veit2016residual}; \nm{gromov2025unreasonable}; \nm{men2025shortgpt}; \nm{lad2024remarkable} \tabularnewline\addlinespace
\raggedright Inference flows of untrained runs go to the collapse region or the noise region depending on the initialization scale (Fig.~\ref{fig:diagram}(a))
& \raggedright Random Transformers exhibit rank collapse or entropy collapse depending on the initialization scale of attention; signal propagation in random deep networks lies in an ordered or a chaotic phase depending on the initialization scale; networks initialized near the boundary between the two are most easily trained
& \raggedright \nm{dong2021attention}; \nm{noci2022signal}; \nm{giorlandino2026two}; \nm{poole2016exponential}; \nm{schoenholz2017deep}; \nm{bahri2020statistical} \tabularnewline\addlinespace
\raggedright The $u \to \infty$ limit of the drifting phase
& \raggedright Rank collapse; merging of token representations into a single cluster in self-attention dynamics; neural collapse with documents as classes
& \raggedright \nm{dong2021attention}; \nm{noci2022signal}; \nm{geshkovski2023emergence}; \nm{geshkovski2023mathematical}; \nm{papyan2020prevalence} \tabularnewline\addlinespace
\raggedright During training, the inference flow enters the working phase from the drifting phase (Fig.~\ref{fig:transition}); $\dC$ first drops sharply, then recovers, and falls back late in training, and $E$ rises early (Fig.~\ref{fig:A-pythia})
& \raggedright During pretraining, the effective rank of representations first drops sharply, then recovers, and later decreases as the representations are compressed toward a few principal directions; during long loss plateaus, the hidden states of different tokens are nearly parallel; probe accuracy is established early in pretraining
& \raggedright \nm{li2025tracing}; \nm{gopalani2025what}; \nm{liu2021probing} \tabularnewline
\bottomrule
\end{tabular}
\end{table*}
\endgroup

\begin{figure}[htbp]
\includegraphics[width=\columnwidth]{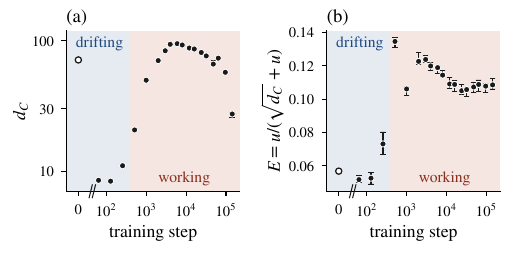}
\caption{Changes in the Pythia-1.4B inference flow over training agree with existing observations: (a) $\dC$ first drops sharply, then recovers, and falls back late, corresponding to the sharp drop, recovery, and late decrease of the effective rank during pretraining~\cite{li2025tracing}; (b) $E=u/(\sqrt{\dC}+u)$ rises early in training, then falls back slightly and levels off, corresponding to probe accuracy being established early in pretraining~\cite{liu2021probing}. The horizontal axis is the training step, on a logarithmic scale with a break between steps 0 and 64. At each checkpoint, we take for each document the late-half medians of $u$ and $\dC$ (relative depth at least 1/2), compute $E$ from them, and take the median over the 96 documents; error bars are 16\%--84\% intervals from a bootstrap over documents and in (a) are mostly smaller than the points; the open point is step 0 (initialization). Background shading marks the two phases as determined from Fig.~\ref{fig:transition}(a): light blue for the drifting phase and light red for the working phase, with the boundary between steps 256 and 512.}
\label{fig:A-pythia}
\end{figure}

\section{Data and models}\label{app:data}

\subsection{Corpus}\label{app:data:corpus}

The corpus consists of 96 coherent long documents from three categories of text, 32 each, namely books (PG-19~\cite{Rae2020Compressive}), papers (arXiv~\cite{Cohan2018Discourse}), and government reports (GovReport~\cite{Huang2021GovReport}). We tokenize each document with each model's own tokenizer, and its first 16\,384 tokens (a length all 96 documents reach) form one input sequence; for Pythia-1.4B, whose context length is 2048, we take the first 2048. All models and all runs use the same set of documents.

\subsection{Trained models}\label{app:data:trained}

The 8 trained models, listed in Table~\ref{tab:models}, come from 5 architecture families (Qwen2.5~\cite{qwen2025qwen25}, Qwen3.5~\cite{qwen35,yang2025qwen3}, Mistral~\cite{jiang2023mistral}, Falcon3~\cite{falcon3}, and Granite~\cite{granite2024granite}) and are all loaded offline at the revisions given in the table (the two Qwen3.5 models are local snapshots of those revisions). All runs use PyTorch~\cite{paszke2019pytorch} 2.6.0, with Transformers~\cite{wolf2020transformers} 5.2 for Qwen3.5 and 4.56.1 for the others. Seven of these are pretrained-only base models; Qwen3.5-9B is the official post-trained version.

\begin{table*}
\caption{The 8 trained models. Here $L$ is the number of blocks, $D$ the residual-stream width, and Revision the Hugging Face commit hash.}
\label{tab:models}
\begin{ruledtabular}
\begin{tabular}{llrrl}
Model & Identifier & $L$ & $D$ & Revision \\
\hline
Qwen2.5-1.5B & \texttt{Qwen/Qwen2.5-1.5B} & 28 & 1536 & \texttt{\scriptsize 8faed761d45a263340a0528343f099c05c9a4323} \\
Qwen2.5-7B & \texttt{Qwen/Qwen2.5-7B} & 28 & 3584 & \texttt{\scriptsize d149729398750b98c0af14eb82c78cfe92750796} \\
Qwen2.5-14B & \texttt{Qwen/Qwen2.5-14B} & 48 & 5120 & \texttt{\scriptsize 97e1e76335b7017d8f67c08a19d103c0504298c9} \\
Mistral-7B-v0.3 & \texttt{mistralai/Mistral-7B-v0.3} & 32 & 4096 & \texttt{\scriptsize caa1feb0e54d415e2df31207e5f4e273e33509b1} \\
Falcon3-7B & \texttt{tiiuae/Falcon3-7B-Base} & 28 & 3072 & \texttt{\scriptsize bf3d7ed586cb22a921520e2d681a9d3d7642cde8} \\
Granite-3.1-8B & \texttt{ibm-granite/granite-3.1-8b-base} & 40 & 4096 & \texttt{\scriptsize 39975ba909950a1bad8fa1cb6981d4d048b75553} \\
Qwen3.5-4B & \texttt{Qwen/Qwen3.5-4B-Base} & 32 & 2560 & \texttt{\scriptsize 1001bb4d826a52d1f399e183466143f4da7b741b} \\
Qwen3.5-9B & \texttt{Qwen/Qwen3.5-9B} & 32 & 4096 & \texttt{\scriptsize c202236235762e1c871ad0ccb60c8ee5ba337b9a} \\
\end{tabular}
\end{ruledtabular}
\end{table*}

\subsection{Untrained runs}\label{app:data:untrained}

The untrained runs use the architectures of the trained models, replacing only the parameters inside the Transformer blocks with random samples; the input embedding and the final norm keep their trained values. The initialization comes in two families, with the initialization scale $\sigma$~\cite{glorot2010understanding} as the control parameter. In family F, the matrix parameters inside the blocks are drawn independently from $\mathcal N(0,\sigma^2)$, the norm gains are 1, and the biases are 0; $\sigma=0.02$ is the default initialization of these architectures in Hugging Face (for Qwen3.5, we obtain this default by calling the model's own initialization function). In family G, all parameters inside the blocks, including the norm gains, are drawn independently from $\mathcal N(0,\sigma^2)$. The two families use the same random numbers, with the seed of each block determined by the run seed and the block index; when the model's own initialization function is called, only the run seed is set. The runs are listed in Table~\ref{tab:untrained}.

\begin{table}
\caption{The 156 untrained runs. Families F and G at $\sigma=0.02$ and family G at $\sigma=1$ each comprise 8 architectures $\times$ 3 seeds (41, 42, 43); each remaining $\sigma$ has 6 runs, namely Qwen2.5-1.5B (seeds 41, 42, 43) and Falcon3-7B, Mistral-7B-v0.3, and Granite-3.1-8B (seed 41).}
\label{tab:untrained}
\begin{ruledtabular}
\begin{tabular}{lp{0.33\columnwidth}r}
Family & $\sigma$ & Runs ($\sigma$ values $\times$ runs per $\sigma$) \\
\hline
F & \raggedright 0.02 & $1\times24=24$ \tabularnewline
F & \raggedright 0.005, 0.01, 0.014, 0.028, 0.04, 0.06, 0.1, 0.2 & $8\times6=48$ \tabularnewline
G & \raggedright 0.02, 1 & $2\times24=48$ \tabularnewline
G & \raggedright 0.035, 0.06, 0.1, 0.17, 0.3, 0.5 & $6\times6=36$ \tabularnewline
\hline
Total & & 156 \\
\end{tabular}
\end{ruledtabular}
\end{table}

\subsection{Pythia training checkpoints}\label{app:data:pythia}

To use training as the control parameter, we take Pythia-1.4B~\cite{biderman2023pythia} (\texttt{EleutherAI/pythia-1.4b}, $L=24$, $D=2048$). Its release includes 154 checkpoints of a single training run from initialization to the end of training (revision \texttt{step}$N$), namely step 0, the power-of-2 steps from 1 to 512, and one every 1000 steps thereafter; there is no checkpoint between steps 256 and 512. We take 19 of them, spaced roughly uniformly in the logarithm of the step, namely steps 0, 64, 128, 256, 512, 1000, 2000, 3000, 4000, 6000, 8000, 12\,000, 16\,000, 24\,000, 32\,000, 48\,000, 64\,000, 96\,000, and 143\,000. Step 0 is the untrained network and is not counted among the 156 untrained runs; step 143\,000 is the released version.
\subsection{From the residual stream to collective coordinates}\label{app:data:measurement}

Measurements use the residual stream at the outputs of blocks 1 through $L-1$ (Sec.~\ref{sec:coordinates:measurement}). The forward propagation is run in bfloat16~\cite{kalamkar2019study}, and statistics are computed in float64.

\emph{Token cleaning.} For each block and document, we remove the first 16 positions and the last position, and then any remaining token whose representation norm exceeds 8 times the median norm over the remaining positions. This cleaning removes the attention sink at the start of the sequence~\cite{xiao2024efficient} and the massive activations at individual positions~\cite{sun2024massive}. The number of retained tokens of document $d$ at a given block is denoted $n_d$.

\emph{Reference.} For document $d$, the reference is the mean of the representations of all retained tokens of the other 95 documents,
\begin{equation}
\mu_{\mathrm{ref}}=\frac{\sum_{d'\neq d}n_{d'}\,\mu_{d'}}{\sum_{d'\neq d}n_{d'}} .
\label{eq:muref}
\end{equation}
The document dimension $\dC$ depends only on $\tr C$ and $\tr C^2$ and is independent of the reference; $u$ requires in addition only $|\delta|^2$, computed from the $\mu_d$ and $n_d$ of each document via Eq.~\eqref{eq:muref}. Hence, for each document and each layer, only $n_d$, $\tr C$, $\tr C^2$, and $|\delta|^2$ need to be stored.

\emph{Document median, relative depth, and late half.} At each layer, the median over the 96 documents gives a point in $(\log_{10}u,\log_{10}\dC)$; connected layer by layer, these points form the trajectory of a run. Models with different numbers of blocks are compared using the relative depth $r=(t-1)/(L-2)$ of block $t$ (Sec.~\ref{sec:coordinates:measurement}). For the cross-model median path (Fig.~\ref{fig:coordinates}(b)--(d)), we linearly interpolate each model's $\log_{10}u$ and $\log_{10}\dC$ in $r$ onto 41 equally spaced grid points and take the median over models at each grid point. The late half refers to the blocks with $r\ge1/2$; the median of a network's document-median trajectory over the late half is called its late-half median position, and $\Pstar$ is the median of the late-half median positions of the 8 trained models (Sec.~\ref{sec:diagram}).

\subsection{Distinctions within the drifting phase}\label{app:data:fates}

\begin{figure}
\includegraphics[width=\columnwidth]{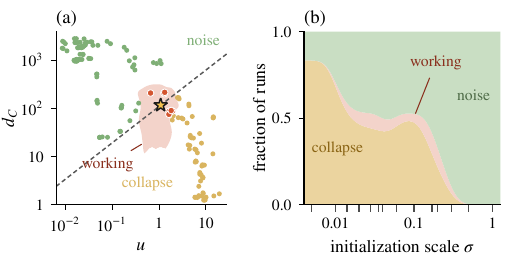}
\caption{Fates of the 156 untrained runs: those with small initialization scales go to the collapse region, and those with large scales to the noise region. (a) Each point is the late-half median position of one run, colored by fate: yellow for the collapse region, green for the noise region, and red for the 8 runs in the working-phase class. Light-red shading marks the working phase; the dashed line is the dividing line through $\Pstar$ (star) along the restoring direction of $J$ of the untrained group; both axes are logarithmic. (b) Fraction of runs with each of the three fates versus the initialization scale $\sigma$, in lighter tints of the three colors in (a); runs from families F and G are pooled with equal weight, and the fractions are smoothed in $\log_{10}\sigma$ with a Gaussian kernel of standard deviation 0.15; short ticks at the bottom mark the sampled values of $\sigma$. The collapse-region fraction drops twice, corresponding to the switch from the collapse region to the noise region in family F ($\sigma$ between 0.014 and 0.04) and in family G ($\sigma$ between 0.17 and 0.3).}
\label{fig:fates}
\end{figure}

Figure~\ref{fig:fates} shows the fates of the 156 untrained runs. The deep-layer points of each run (96 documents, late-half layers) are divided into three classes. Points within the working phase (Appendix~\ref{app:sensitivity:phase}) form one class; the remaining points belong to the collapse region (the side with large $u$ and small $\dC$) or the noise region, according to which side of the dividing line they lie on. The dividing line passes through $\Pstar$ along the restoring direction of the linear response matrix $J$ of the untrained group (the eigendirection with a negative eigenvalue; Fig.~\ref{fig:phases}(a)). Each run is assigned to the class holding the largest fraction of its points. The working-phase, collapse-region, and noise-region classes contain 8, 58, and 90 runs, respectively. Inference flows with small initialization scales mostly go to the collapse region, and those with large scales go to the noise region (Fig.~\ref{fig:fates}(b)). The transition occurs at different scales in the two families, between $\sigma=0.014$ and 0.04 for family F and between 0.17 and 0.3 for family G.

\begin{figure}
\includegraphics[width=\columnwidth]{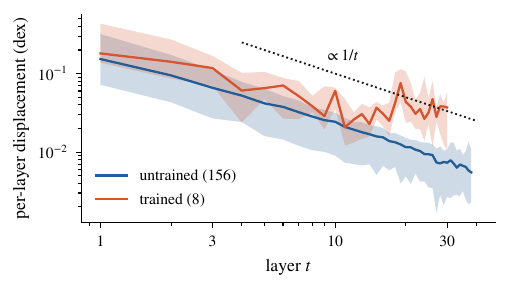}
\caption{Inference flows of untrained runs slow down progressively in the deep layers. Per-layer displacement of the document-median trajectory in $(\log_{10}u,\log_{10}\dC)$ versus the layer index $t$, on log-log axes. Blue: the 156 untrained runs; red: the 8 trained models. Lines are within-group medians and bands within-group 16\%--84\% intervals, drawn only for layers reached by at least 20 runs or 3 models. The dotted line is a reference with slope $-1$ ($\propto1/t$).}
\label{fig:slow}
\end{figure}

Inference flows of untrained runs slow down progressively in the deep layers (Fig.~\ref{fig:slow}); the displacement per layer decreases roughly as $1/t$, where $t$ is the layer index. The late-half log-log slope of displacement versus layer index has a median of $-0.93$ over the 156 runs and is negative for 97\% of them. Under a $1/t$ extrapolation, doubling the depth moves the flow only about 0.17~dex further (median over runs). The displacement of the trained models no longer decreases in the deep layers and is dominated by fluctuations.

We measure dispersion by the root-mean-square distance in $(\log_{10}u,\log_{10}\dC)$ of the late-half median positions from the within-group mean position. The dispersion is 0.085~dex for the 8 trained models. For each set of 24 untrained runs (8 architectures, 3 seeds) sharing a family and $\sigma$, it is 0.28~dex (family G, $\sigma=1$), 1.25~dex (family G, $\sigma=0.02$), and 1.38~dex (family F, $\sigma=0.02$).

\section{Collective coordinates and linear readout}\label{app:coordinates}\label{app:coordinates:readout}

Blocks read the residual stream through linear projections~\cite{elhage2021mathematical,gurnee2026verbalizable}, so $(u,\dC)$ can be interpreted through a linear readout. Let $y_i=h_i-\mu_{\mathrm{ref}}$ be the representation of the $i$th token in a document relative to the reference ($i=1,\dots,n_d$, where $n_d$ is the number of tokens retained after cleaning; Appendix~\ref{app:data:measurement}), and let $E$ be the ratio of the mean inner product between pairs of token representations in the same document (the two-point correlator) to their mean squared norm~\cite{song2025bridging}; as $n_d\to\infty$,
\begin{equation}
E=\frac{\sum_{i\ne i'}y_i\cdot y_{i'}}{(n_d-1)\sum_i|y_i|^2}
\to\frac{|\delta|^2}{\tr C+|\delta|^2}=\frac{u}{\sqrt{\dC}+u}.
\label{eq:B-E}
\end{equation}
When the norms of the token representations (relative to the reference) are similar, $E$ is approximately the average cosine similarity between two token representations in the same document~\cite{song2025bridging}.

\begin{figure}[b]
\includegraphics[width=\columnwidth]{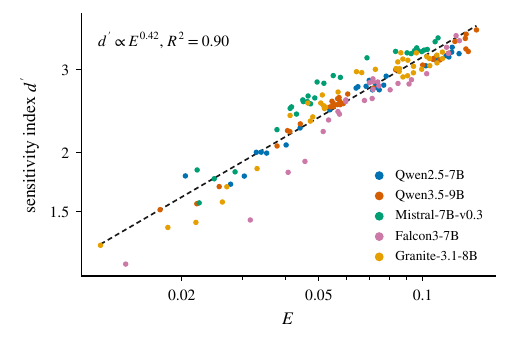}
\caption{The document identification accuracy rises with $E$. Each point is one block of one trained model (5 models, 155 blocks in total), colored by model. The horizontal axis is the median of $E$ over the 96 documents at that block, and the vertical axis is the sensitivity index $d'$ converted from the test accuracy of the 96-class logistic regression, both on log scales. The dashed line is the least-squares power-law fit to all blocks in log-log coordinates; the fitted exponent and $R^2$ are given at upper left.}
\label{fig:readout}
\end{figure}

The document identification accuracy is the accuracy with which a single token's representation alone identifies which of the 96 documents the token comes from. We measure it at the outputs of blocks $1$ to $L-1$ of five trained models (Qwen2.5-7B, Qwen3.5-9B, Mistral-7B-v0.3, Falcon3-7B, Granite-3.1-8B), 155 blocks in total. The 16\,384 tokens of each document are cut into segments of 1024 tokens; the first segment is discarded, and the remaining 15 are randomly split 9/3/3 into training, validation, and test segments. Each document's split is the same for all blocks. Within each segment, we take every 128th token. We fit a 96-class multinomial logistic regression on the first $K$ principal components (each standardized) of the training token representations, selecting $K\in\{128,256,512\}$ and the L2 regularization coefficient from $\{10^{-1},\dots,10^{-4}\}$ by validation accuracy; we then refit on the training and validation segments and report the test accuracy $A$. The test accuracy $A$ is converted to the sensitivity index $d'$ through the equal-variance Gaussian model $A(d')=\int\phi(z)\,\Phi(z+d')^{95}\,dz$~\cite{hacker1979revised}, where $\phi$ and $\Phi$ are the standard normal density and distribution function. The document identification accuracy rises with $E$ (Fig.~\ref{fig:readout}), with $d'\propto E^{0.42}$ and $R^2=0.90$.

\section{Linear map and contraction rate}\label{app:linear}

This appendix derives $M$ in Eq.~\eqref{eq:propagator} from Eq.~\eqref{eq:langevin}, shows why $\rho<1$ is equivalent to $J$ being a stable node for the $J$ and $\alpha$ of the two groups, and presents an ablation with per-network estimation.

\subsection{Map}\label{app:linear:map}

Substituting Eq.~\eqref{eq:linear} into Eq.~\eqref{eq:langevin}, with $x$ measured relative to $\Pstar$, and using $x_{t+1}=x_t+v_{t+1}$, we obtain
\begin{equation}
z_{t+1}=M z_t+\begin{pmatrix}a+\eta_t\\ a+\eta_t\end{pmatrix},\qquad z_t=\begin{pmatrix}x_t\\ v_t\end{pmatrix},
\label{eq:D-affine}
\end{equation}
where $M$ is the matrix in Eq.~\eqref{eq:propagator}. The constant term $a$ and the fluctuation both enter additively. When one inference flow is subtracted from another, or the average of the flows from one inference flow, $a$ cancels, the fluctuation acts only as a source term, and the deviation propagates layer by layer under the same $M$. The spectral radius $\rho$ characterizes how inference flows contract or spread apart relative to one another and to the average of the flows. The effective force $F$ is evaluated at the starting point $x_t$ of the transition (Appendix~\ref{app:estimation}), so $M$ defines an explicit iteration; Fig.~\ref{fig:F-settings} shows the results with $F$ evaluated at the midpoint or end point instead.

\subsection{Criterion}\label{app:linear:criterion}

Let $w$ be an eigenvector of $J$ with eigenvalue $\lambda$. Then $M$ maps the subspace $\{(p\,w,\ q\,w)\}$ into itself, and its eigenvalues $\mu$ on this subspace satisfy
\begin{equation}
\chi(\mu)\equiv\mu^2-(1+\lambda+\alpha)\,\mu+\alpha=0 .
\label{eq:D-charpoly}
\end{equation}
Each of the two eigenvalues of $J$ gives two roots, which together are the four eigenvalues of $M$, and the spectral radius $\rho$ of $M$ is the largest of their moduli. The deviation scales by roughly a factor $\rho$ per layer and contracts when $\rho<1$. By the Jury criterion~\cite{jury1962simplified}, both roots lie inside the unit circle if and only if $|\alpha|<1$ and $-2(1+\alpha)<\lambda<0$. For a saddle, one eigenvalue of $J$ satisfies $\lambda>0$, and $\rho>1$ for any $\alpha$. For the $J$ and $\alpha$ of the two groups, $\rho<1$ if and only if $J$ is a stable node.

\subsection{Ablation with per-network estimation}\label{app:linear:quenched}

Section~\ref{sec:phases} pools the transitions of a group of networks and fits a single $\alpha$ and a single $J$. For the trained group, we instead estimate $J$ for each model separately. We set $\alpha$ to the group value and linearize each model at its own late-half median position, with the fit range unchanged and everything else as in Appendix~\ref{app:estimation}. The median of $-\ln\rho$ over the 8 models is 0.027, and all 8 values are positive. Most untrained runs do not pass through the working phase; per-run estimation could use only the few runs that do and would be biased, so we apply per-network estimation only to the trained group.

\section{Estimation methods}\label{app:estimation}\label{app:estimation:cells}\label{app:estimation:alpha}\label{app:estimation:J}\label{app:estimation:ref}

Below, a network means an untrained run, a trained model, or a Pythia checkpoint; the coordinates are $x=(\log_{10}u,\log_{10}\dC)$, and displacements, forces, and $J$ are all per layer. For the two groups, the linearization point is $\Pstar=(1.08,117)$ and the fit range is $0.3\le u\le3$, $30\le\dC\le400$; for Pythia, the linearization point is its representative point $\Pstar_{\rm Py}=(0.63,27)$, the late-half median position of the final checkpoint, and the fit range spans the same multiples of $\Pstar_{\rm Py}$, namely $0.17\le u\le1.75$, $7.0\le\dC\le93$.

\subsection{Joint estimation for a group of networks (Algorithm 1)}\label{app:estimation:joint}

Algorithm 1 gives the values of $\alpha$, $J$, and $-\ln\rho$ shown in Figs.~\ref{fig:phases} and~\ref{fig:transition}.
\par\noindent\begin{minipage}{\linewidth}
\begin{framed}\begin{algorithmic}[1]
\Require a group of networks, 96 documents, linearization point $\Pstar$ ($\Pstar_{\rm Py}$ for Pythia)
\State for each network and each document, every three consecutive layers $(t-1,t,t+1)$ give one transition $(x_t,\,v_t=x_t-x_{t-1},\,v_{t+1}=x_{t+1}-x_t)$
\State $\hat\alpha\gets$ regression slope of $v_{t+1}$ on $v_t$ after demeaning within classes (network, 0.1~dex cell containing $x_t$), shared by the two coordinates
\State $f_t\gets v_{t+1}-\hat\alpha\,v_t$
\State divide the fit range into 0.1~dex cells (origin at $\Pstar$); $F_k\gets$ median of $f_t$ in cell $k$ (at least 20 transitions)
\State $w_k\gets 1/\sum_i\mathrm{Var}(F_{k,i})$ ($i$ runs over the two coordinates), with variances from 50 preliminary resamples (Algorithm 2); only cells appearing in at least 10 of them are used
\State weighted least-squares fit of $F_k\approx a+J\,(c_k-\Pstar)$, with $c_k$ the cell center, giving $\hat J$
\State $\rho\gets$ spectral radius of $M(\hat J,\hat\alpha)$ in Eq.~\eqref{eq:propagator}
\Ensure $\hat\alpha$, $\hat J$, $-\ln\rho$
\end{algorithmic}\end{framed}
\end{minipage}\par\medskip

\subsection{Bootstrap (Algorithm 2)}\label{app:estimation:bootstrap}

\par\noindent\begin{minipage}{\linewidth}
\begin{framed}\begin{algorithmic}[1]
\For{$m=1,\dots,200$}
  \State resample the networks of the group with replacement (skipped when there is only one network)
  \State resample the 96 documents with replacement, shared by all networks
  \State redo lines 1--4 and 6--7 of Algorithm 1 ($w_k$ unchanged)
\EndFor
\Ensure point estimate $\theta$ taken from the full sample; error $\theta^{+(q_{84}-\theta)}_{-(\theta-q_{16})}$, where $q_{16}$ and $q_{84}$ are the 16\% and 84\% quantiles of the 200 resamples
\end{algorithmic}\end{framed}
\end{minipage}\par\medskip

\subsection{Force-field map and Péclet number (Algorithm 3)}\label{app:estimation:peclet}

Algorithm 3 gives the arrows and background shading in Fig.~\ref{fig:phases}.
\par\noindent\begin{minipage}{\linewidth}
\begin{framed}\begin{algorithmic}[1]
\Require $f_t$ from Algorithm 1; 0.1~dex cells within the fit range, on a fixed grid with edges at multiples of 0.1~dex in $(\log_{10}u,\log_{10}\dC)$
\State $\tilde F_k\gets$ median of $f_t$ in the cell, taken first within each network and then across networks (at least 5 transitions per network in the cell, at least 3 networks)
\State $\eta_t\gets f_t-\tilde F_k$; $s_{k,i}\gets1.4826\times\mathrm{median}\,|\eta_{t,i}-\mathrm{median}\,\eta_{t,i}|$ (median absolute deviation~\cite{hampel1974influence,rousseeuw1993alternatives} along the $i$th coordinate within the cell; the factor 1.4826 makes it equal to the standard deviation for a normal distribution)
\State compute $\mathrm{Pe}_k$ according to Eq.~\eqref{eq:peclet}
\end{algorithmic}\end{framed}
\end{minipage}\par\medskip
The median of $\mathrm{Pe}$ is 1.99 in the untrained group and 0.49 in the trained group.

\section{Sensitivity, Pythia binning, phase criterion, and delineation of the phase regions}\label{app:sensitivity}

\subsection{Sensitivity of the results of Sec.~\ref{sec:phases}}\label{app:sensitivity:ref}

\begin{figure*}
\includegraphics[width=\textwidth]{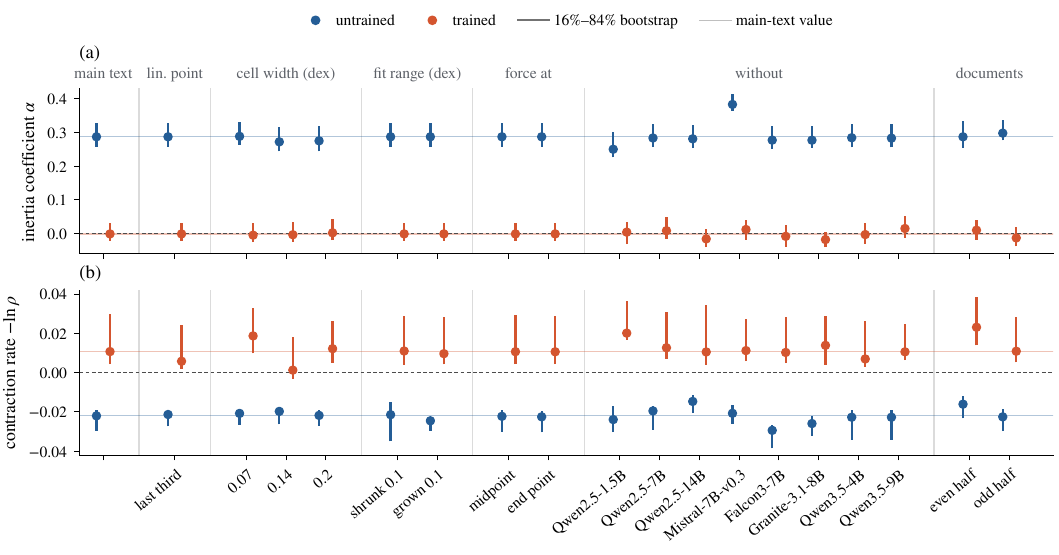}
\caption{(a) Inertia coefficient $\alpha$ and (b) contraction rate $-\ln\rho$ of the two groups of Sec.~\ref{sec:phases} under different analysis settings. Each column changes only one setting (Table~\ref{tab:F-settings}); the columns are arranged by setting category, with category names marked above (a). Blue is the untrained group and red the trained group; points are estimates, and vertical bars are 16\%--84\% intervals from 200 bootstrap resamples; light horizontal lines are the main-text values, and the dashed line is 0.}
\label{fig:F-settings}
\end{figure*}

Figure~\ref{fig:F-settings} gives the estimates of $\alpha$ and $-\ln\rho$ for the two groups of Sec.~\ref{sec:phases} under different analysis settings. Each column of the figure changes only one setting and corresponds to one row of Table~\ref{tab:F-settings}, with everything else as in Appendix~\ref{app:estimation}; all errors are 16\%--84\% intervals from 200 bootstrap resamples. Under every setting, the untrained group has $\alpha>0$ and $-\ln\rho<0$, and the trained group has $\alpha$ close to 0 and $-\ln\rho>0$, as in the main text; the results of Sec.~\ref{sec:phases} are robust to these choices.

\begin{table}
\caption{Setting changed in each column of Fig.~\ref{fig:F-settings}; rows from top to bottom correspond to the figure's columns from left to right. Unlisted settings follow Appendix~\ref{app:estimation} (all networks, 96 documents, fit range $0.3\le u\le3$, $30\le\dC\le400$).}
\label{tab:F-settings}
\begin{ruledtabular}
\begin{tabular}{p{0.3\columnwidth}p{0.58\columnwidth}}
\raggedright Column in the figure & \raggedright Setting changed \tabularnewline
\hline
\raggedright main text & \raggedright unchanged \tabularnewline
\raggedright lin. point: last third & \raggedright late half replaced by the last third ($r\ge2/3$), linearization point moved to $(1.28,112)$ \tabularnewline
\raggedright cell width: 0.07 & \raggedright cell width changed from 0.1~dex to 0.07~dex \tabularnewline
\raggedright cell width: 0.14 & \raggedright changed to 0.14~dex \tabularnewline
\raggedright cell width: 0.2 & \raggedright changed to 0.2~dex \tabularnewline
\raggedright fit range: shrunk 0.1 & \raggedright each edge of the fit range moved inward by 0.1~dex \tabularnewline
\raggedright fit range: grown 0.1 & \raggedright each edge of the fit range moved outward by 0.1~dex \tabularnewline
\raggedright force at: midpoint & \raggedright estimation algorithm unchanged; in computing $\rho$, the linear term of the effective force in Eq.~\eqref{eq:propagator} is evaluated at $(x_t+x_{t+1})/2$ instead of $x_t$ \tabularnewline
\raggedright force at: end point & \raggedright as above, but at $x_{t+1}$ \tabularnewline
\raggedright without: X (8 columns) & \raggedright networks of architecture X (its untrained runs and trained models) removed from both groups at once; X runs over the 8 architectures \tabularnewline
\raggedright documents: even half & \raggedright only the 48 even-numbered documents used \tabularnewline
\raggedright documents: odd half & \raggedright only the 48 odd-numbered documents used \tabularnewline
\end{tabular}
\end{ruledtabular}
\end{table}

\begin{figure}
\includegraphics[width=\columnwidth]{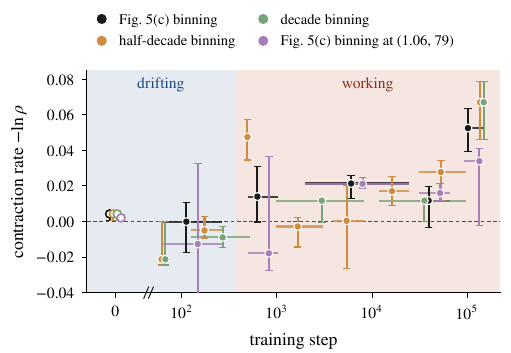}
\caption{Contraction rate $-\ln\rho$ of Pythia-1.4B from per-bin joint fits under different binnings and linearization points. The horizontal axis is the training step, on a logarithmic scale, broken between steps 0 and 64. Black is the binning of Fig.~\ref{fig:transition}(c), orange is half-decade binning, and green is decade binning, all linearized at $\Pstar_{\rm Py}=(0.63,27)$; purple is the black binning linearized at $(1.06,79)$, where the 64--256-step bin has only 9 cells near that point. Points are per-bin fitted values, open points mark the step-0 bin, and points of different colors in the same bin are slightly offset horizontally; vertical bars are 16\%--84\% bootstrap intervals, clipped at the vertical-axis limits; horizontal bars span the step range of the bin, and bins containing only one checkpoint have none; the dashed line is 0. The blue and red backgrounds mark the drifting phase and working phase, determined by the fraction of deep-layer states falling in the working phase (Fig.~\ref{fig:transition}(a)).}
\label{fig:F-bins}
\end{figure}

\subsection{Pythia binning}\label{app:sensitivity:bins}

The binning of Fig.~\ref{fig:transition}(c) puts step 0 in its own bin, followed by five bins of 64--256, 512--1000, 2000--24\,000, 32\,000--64\,000, and 96\,000--143\,000 steps; each of these five bins has at least 2 checkpoints and at least 20 cells near the linearization point $\Pstar_{\rm Py}$. Figure~\ref{fig:F-bins} compares this binning with half-decade binning (boundaries at $10^{k/2}$ steps) and decade binning (boundaries at $10^{k}$ steps), both of which also put step 0 in its own bin; we also fit the binning of Fig.~\ref{fig:transition}(c) at the linearization point $(1.06,79)$, the median of the late-half median positions of the checkpoints from step 6000 on. Under all binnings and linearization points, $-\ln\rho$ is about 0 in the drifting phase and mostly positive after entry into the working phase, with no jump, consistent with the conclusion of Sec.~\ref{sec:transition}.

\begin{figure}
\includegraphics[width=\columnwidth]{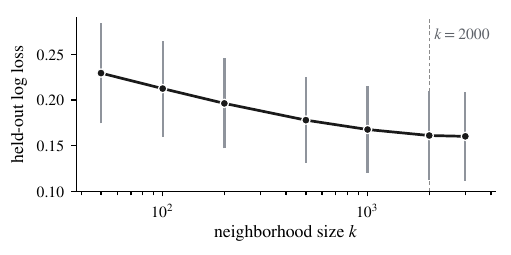}
\caption{Held-out log loss versus neighborhood size $k$ in leave-one-architecture-out cross-validation (two groups weighted equally). Points are averages over the 8 architectures, and vertical bars are standard errors across architectures; the dashed line marks the chosen $k=2000$.}
\label{fig:F-k}
\end{figure}

\subsection{Phase criterion}\label{app:sensitivity:criterion}

Section~\ref{sec:transition} determines the two phases by the fraction of deep-layer states in the working phase. A checkpoint is assigned to the working phase if more than half of its deep-layer points (96 documents, each late-half layer) fall in the working phase (Fig.~\ref{fig:transition}(a)); points in blank areas count toward the total but not toward the working phase. The per-checkpoint $\alpha$ also drops sharply between steps 256 and 512 (Fig.~\ref{fig:transition}(b)). Step 0 is an exception and is assigned to the drifting phase. Although 89\% of its deep-layer points fall in the working phase, the cosine between successive per-layer displacements of its document-median trajectory, averaged over the late half, is 0.99; the inference flow is still drifting in the same direction.

\subsection{Lever rule for the inertia coefficient}\label{app:sensitivity:lever}

In a two-phase coexistence region, an observable is the average of its values in the two pure phases weighted by the phase fractions (the lever rule~\cite{callen1985thermodynamics}). We model the inertia coefficient of checkpoint $s$ as $\alpha(s)=\varphi(s)\,\alpha_W+[1-\varphi(s)]\,\alpha_D$, where $\varphi(s)$ is the working-phase fraction of Fig.~\ref{fig:transition}(a); $\alpha_W$ and $\alpha_D$ are the values in the working and drifting phases. A least-squares fit over the 18 checkpoints with $s>0$ gives $\alpha_W=-0.035^{+0.018}_{-0.017}$ and $\alpha_D=0.538^{+0.056}_{-0.054}$ (16\%--84\% intervals from resampling the checkpoints with replacement within each phase, 2000 times). Step 0 is excluded: its deep-layer states fall in the working phase ($\varphi=0.89$), yet $\alpha$ keeps its drifting value and $-\ln\rho$ is about 0; the inference flow is still drifting and only passes through the working phase at finite depth (Sec.~\ref{sec:transition}).

\subsection{Delineation of the phase regions}\label{app:sensitivity:phase}

The background shading and the coexistence band of Fig.~\ref{fig:diagram} are obtained from all deep-layer points. Each network gives one point per document for each late-half layer, 252\,288 points in total, 12\,864 of them from the trained group; points of the trained group are labeled working phase, and those of the untrained group drifting phase. Networks in the same group carry equal weight, divided equally among their points, and the two groups carry equal total weight. The working-phase fraction $p$ at any position is the weight fraction of the trained group among the $k$ nearest points. The region with $p>0.5$ is the working phase, and that with $p$ between 0.16 and 0.84 is the coexistence band; positions with no point within 0.1~dex (the cell width of Sec.~\ref{sec:phases}) are left unclassified and blank in Fig.~\ref{fig:diagram}. The two red lines in that figure are the $p=0.16$ and $0.84$ contours of the Gaussian-smoothed $p$ (standard deviation 0.03~dex); classification and statistics both use the unsmoothed $p$, and smoothing changes the fraction of each Pythia checkpoint's points falling in the working phase by no more than 0.004.

We choose the neighborhood size $k$ by leave-one-architecture-out cross-validation, holding out all networks of one architecture in turn and predicting the group of each held-out point from the remaining points; we take $k=2000$, where the log loss stops decreasing (Fig.~\ref{fig:F-k}). The entry depth 0.11 in the main text is the median, over the 8 trained models, of the relative depth at which the document-median trajectory first enters the working phase.

\end{document}